\documentclass[10pt,twocolumn,letterpaper]{article}

\usepackage[pagenumbers]{wacv} 

\usepackage{graphicx}
\usepackage{amsmath}
\usepackage{amssymb}
\usepackage{booktabs}
\usepackage{epsfig}
\usepackage{bm}
\usepackage{bbm}
\usepackage{multirow}
\usepackage{xcolor}
\usepackage{algorithm2e} 
\usepackage{dsfont} 
\SetKwInput{KwInput}{Input}                
\SetKwInput{KwOutput}{Output}              
\usepackage[normalem]{ulem}
\usepackage{comment}
\newcommand{\R}{{\mathbb R}}
\newcommand{\Mesh}{\mathcal{M}}
\newcommand{\Points}{\mathcal{P}}

\newcommand{\Camera}{\mathcal{C}}
\newcommand{\Focus}{\mathcal{S}}
\newcommand{\FFocus}{{\tilde{\Focus}}}
\newcommand{\pphi}{{\tilde{\phi}}}

\newcommand*\diff{\mathop{}\!\mathrm{d}}
\DeclareMathOperator*{\argmax}{arg\,max}
\DeclareMathOperator*{\argmin}{arg\,min}

\newcommand{\samethanks}[1][\value{footnote}]{\footnotemark[#1]}

\usepackage[pagebackref,breaklinks,colorlinks]{hyperref}

\usepackage[capitalize]{cleveref}
\crefname{section}{Sec.}{Secs.}
\Crefname{section}{Section}{Sections}
\Crefname{table}{Table}{Tables}
\crefname{table}{Tab.}{Tabs.}

\def\wacvPaperID{244} 
\def\confName{WACV}
\def\confYear{2025}

\begin{document}

\title{A Novel Method to Improve Quality Surface Coverage in Multi-View Capture}

\author{Wei-Lun Huang\thanks{Corresponding author: Wei-Lun Huang (wl.huang@jhu.edu)}\\
Johns Hopkins University\\
Baltimore MD 21218, USA\\
National Institutes of Health\\
Bethesda, MD, USA
\and
Davood Tashayyod\\
Lumo Imaging\\
Rockville, MD, USA
\and
Amir Gandjbakhche\\
National Institutes of Health\\
Bethesda, MD, USA
\and
Michael Kazhdan\thanks{Co-senior authors}\\
Johns Hopkins University\\
Baltimore MD 21218, USA
\and
Mehran Armand\samethanks[2]\\
Johns Hopkins University\\
Baltimore MD 21218, USA
}
\maketitle

\begin{abstract}
  The depth of field of a camera is a limiting factor for applications that require taking images at a short subject-to-camera distance or using a large focal length, such as total body photography, archaeology, and other close-range photogrammetry applications. Furthermore, in multi-view capture, where the target is larger than the camera's field of view, an efficient way to optimize surface coverage captured with quality remains a challenge. Given the 3D mesh of the target object and camera poses, we propose a novel method to derive a focus distance for each camera that optimizes the quality of the covered surface area. We first design an Expectation-Minimization (EM) algorithm to assign points on the mesh uniquely to cameras and then solve for a focus distance for each camera given the associated point set. We further improve the quality surface coverage by proposing a $k$-view algorithm that solves for the points assignment and focus distances by considering multiple views simultaneously. We demonstrate the effectiveness of the proposed method under various simulations for total body photography. The EM and $k$-view algorithms improve the relative cost of the baseline single-view methods by at least $24$\% and $28$\% respectively, corresponding to increasing the in-focus surface area by roughly $1550$ cm$^2$ and $1780$ cm$^2$. We believe the algorithms can be useful in a number of vision applications that require photogrammetric details but are limited by the depth of field.
\end{abstract}

\section{Introduction}
\label{sec:intro}
RGB cameras are widely used to capture the visual information of real-world objects: geometry and texture. Due to optical properties, an RGB camera has limited depth of field -- the distance between the closest and the furthest objects that appear with acceptable sharpness in an image \cite{skuka2022extending}. Acquiring visual details at high resolution, as in total body photography\cite{korotkov2014new}, archaeology \cite{gajski2016applications}, and other close-range photogrammetry applications \cite{nocerino2016experiments}, requires taking images at a short subject-to-camera distance or using a large focal length. In both cases, the variation in depth of the target is larger than the depth of field of the cameras, resulting in blurry image regions (e.g. When a camera is capturing from the lateral side of a patient, the thigh closer to the camera will be in focus while the other thigh would not be \cite{korotkov2014new}.)

Extending the depth of field can be achieved by focal stacking or focal sweeping
\cite{kuthirummal2010flexible,hausler1972method}. In practice, extending the depth of field mostly captures multiple images from a single camera pose. The number of required source images can be determined with depth information \cite{skuka2022extending}. However, in applications where focal stacking or focal sweeping are infeasible, such as having a moving camera during a scan and when scanning time is restricted, we would like to take a single image per camera pose. In such multi-view capture, when the target is larger than the camera's field of view, an efficient way to optimize the quality of surface coverage remains a challenge.

Auto-focus (AF) for single-image capture is well-studied and most modern cameras have hardware support that allows quick lens movements for optimizing image sharpness. 
However, in the context of multi-view capture, AF is greedy (i.e. it does not consider if a given point is already seen by some other camera \cite{abuolaim2018revisiting,abuolaim2020online}) and, more generally, may fail when imaging textureless objects.

View planning solves for a set of views (camera poses) that need to be computed for automated object reconstruction considering quality constraints and efficiency simultaneously \cite{scott2003view}. Common constraints and quality criteria include view overlap, efficiency, limited prior knowledge, sensor frustums, sensor pose constraints, and so on. However, the depth of field is usually not considered when calculating the visual coverage in view planning. Furthermore, view planning is computationally complex (shown to be NP-complete \cite{tarbox1995planning,scott2002performance}).

\begin{figure*}[t]
      \centering
      \includegraphics[width=14cm]{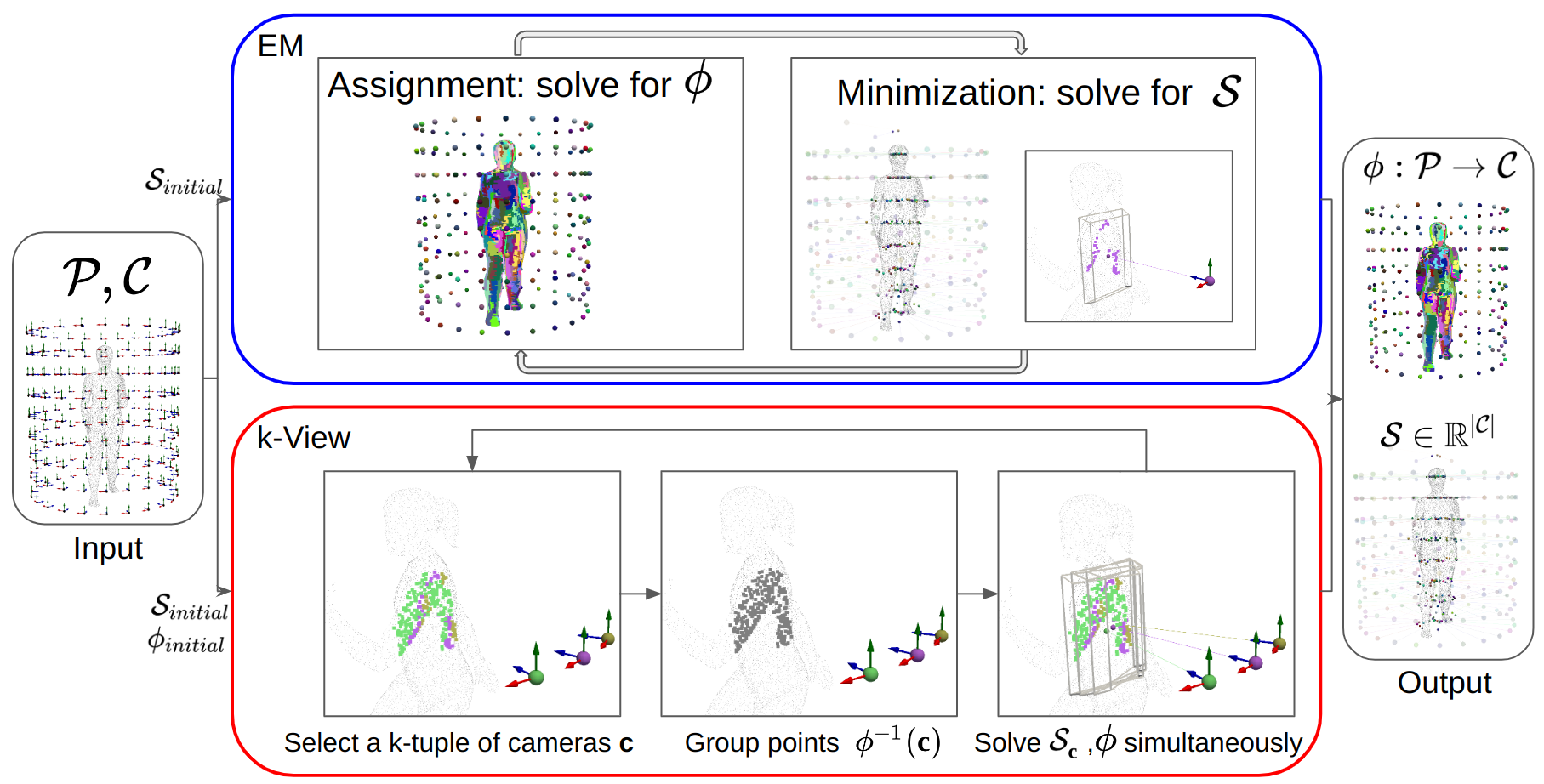}
      \caption{The illustration of the EM and the $k$-view method. The input is the point cloud of the target $\Points$ and cameras $\Camera$. The output is the point assignment function $\phi:\Points\rightarrow\Camera$ and the focus distances $\Focus\in\R^{|\Camera|}$. The camera $c$ and its associated points $\phi^{-1}(c)$ are visualized in the same color.}
      \label{fig:illustration}
   \end{figure*}

Given a 3D mesh of the target object and camera poses, we propose a novel method to select a focus distance for each camera so as to provide high-quality surface coverage. We begin by designing an Expectation-Minimization (EM) algorithm that iteratively assigns points on the mesh to cameras and solves for the focus distance for each camera given the associated point set. As it is greedy and the optimization problem is not convex, the EM algorithm could yield a sub-optimal solution. Inspired by the alpha-beta swap algorithm for graph cut \cite{boykov2001fast}, we improve the surface coverage by proposing a $k$-view algorithm that considers $k$-tuples of cameras simultaneously. \cref{fig:illustration} illustrates the EM and the $k$-view methods. We demonstrate the effectiveness of the proposed method under various simulations.

Our work focuses on the quality surface coverage aggregated from multiple images in a scan. Although 3D reconstruction is a downstream task that can benefit from sharper images in the proposed method, it is beyond the scope of the paper. Furthermore, although we apply the proposed method for total body coverage, our approach readily extends to other contexts in close-range photogrammetry. We will release the source code upon acceptance. Overall, we make three contributions:

\begin{enumerate}
\item
We formulate the multi-view scanning problem when restricted to determining the camera's depth of field and propose the use of a baseline EM method that optimizes focus distances by considering one view at a time.  
\item
We design a novel $k$-view optimization algorithm that improves on the quality of the coverage, compared to the EM method.
\item
We demonstrate the effectiveness of the proposed method in extensive simulations and explore variants of the $k$-view algorithm. 
\end{enumerate}

\section{Related Work}
\label{sec:related work}
\subsection{Extending depth of field}
Extending depth of field by capturing multiple images at different focal planes followed by image fusion is a technique for acquiring an all-in-focus image of a target object \cite{liu2020multi,liu2017multi,kuthirummal2010flexible}. Str{\"o}bel \textit{et al.} \cite{strobel2018automated} proposed an automated device to combine extended depth of field images from multiple views to reconstruct 3D models of pinned insects and other small objects. For extended depth of field in multi-view images, Chowdhury \textit{et al.} \cite{chowdhury2021fixed} proposed to use a fixed-lens camera and calibrated image registration to mitigate artifacts in the fused images due to violation of perspective image formation. However, in multi-focus image fusion applications, the number of source images is usually undetermined without depth information. Recently, Skuka \textit{et al.} \cite{skuka2022extending} proposed a method based on the depth map of the scene for extending the depth field of the imaging systems. Nonetheless, they only focus on a single target scene with the same camera pose across all images. Though the approach extends to multiple camera poses, the computational complexity is exponential in the number of poses for cameras with overlapping fields of view. 

\subsection{View planning }
View planning \cite{scott2003view} solves for a set of views (camera poses) required for automated object reconstruction considering quality constraints and efficiency simultaneously. A large body of research has addressed the problem of view planning for 3D reconstruction, inspection \cite{zeng2020view}, and robotics \cite{vasquez2017view}. The approaches usually focus on finding the next best view: optimizing view selection from the set of available views. Dunn \textit{et al.} \cite{dunn2009next} proposed a hierarchical uncertainty-driven model to select viewpoints based on the model’s covariance structure and appearance and the camera characteristics. Recently, Gu{\'e}don \textit{et al.} \cite{guedon2022scone} proposed using neural networks to predict the occupancy probability of the scene and the visibility of sampled points in the scene given any camera pose, to evaluate the visibility gain for camera poses. Additionally, geometric priors have also been incorporated into view planning for multi-object 3D scanning \cite{fan2016automated} and robotics de-powdering \cite{do2023geometry}. Without geometric priors but given the constraint of the measurement volume, Osi{\'n}ski \textit{et al.} \cite{osinski2022novel} proposed an approach to dynamically design a multi-view stereo camera network for the required resolution and the accuracy of geometric reconstruction. However, view planning is computationally complex. For the applications we focus on in this paper, we assume that calibrated camera poses and a 3D mesh are given, and narrow the task to the optimization of the depth of field for each camera.
\section{Method}
\begin{figure}
  \centering
  \begin{subfigure}{0.45\textwidth}
    \includegraphics[width=\textwidth]{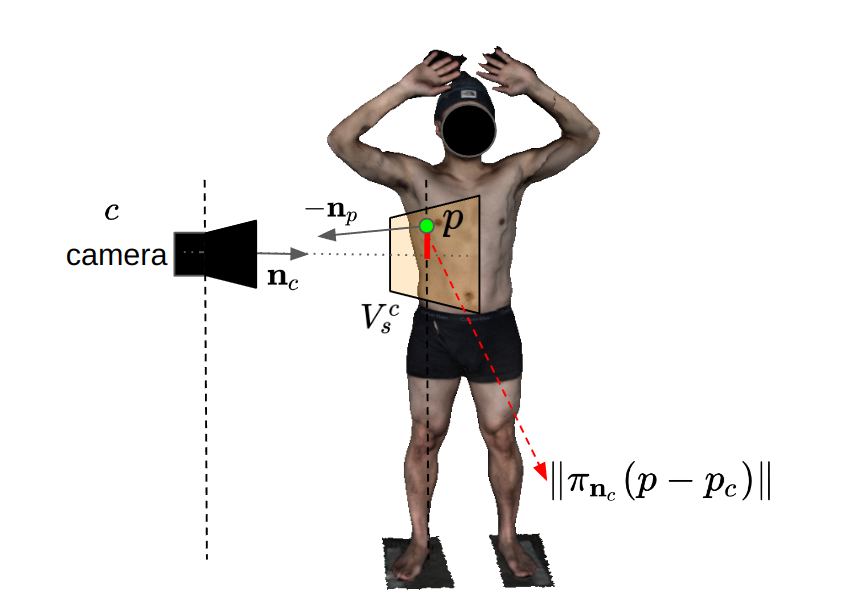}
    \caption{Cost-determining factors}
    \label{fig:criteria}
  \end{subfigure}
  \hfill
  \begin{subfigure}{0.45\textwidth}
    \includegraphics[width=\textwidth]{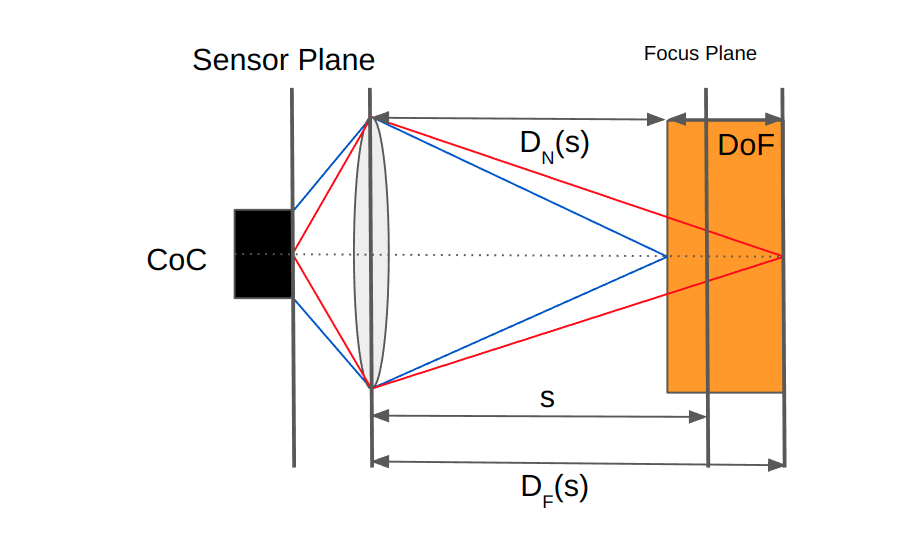}
    \caption{The principle of the depth of field}
    \label{fig:dof}
  \end{subfigure}
  \caption{Visualization for the cost-determining factors and the principle of the depth of field. In (a), the orange frustum ($V^c_s\subset\R^3$) is clipped at the near and far depth of field (DoF) limits. The red line represents the deviation from the optical axis. In (b), $CoC$ is the circle of confusion, $s$ is the focus distance, and $D_N(s)$ and $D_F(s)$ are the near and far depth of field limits.}
  \label{fig:criteria_and_dof}
\end{figure}   
\subsection{Problem Statement}
 Given a mesh $\Mesh$ and camera poses $\Camera\subset\mathrm{SE}(3)$, we would like to solve for an assignment of focus distances $\Focus \in \R^{|\Camera|}$ to cameras that reduces the size of ``poorly'' imaged surface. We formulate this by (1)~assigning a per-camera cost to each point of surface, (2)~defining the cost per-point as the minimum cost over all cameras, and (3)~seeking the focus distances that minimize the integrated cost over all points.
\subsection{Cost Function}
\subsubsection{Focus Distance Cost}
Given a camera $c\in\Camera$ and a focus distance $s\in\R$, we define a pointwise cost function $\kappa_s^c:\Mesh\rightarrow[0,1]$. The function is set to one (the maximum cost) for all surface points $p\in\Mesh$ that are invisible to camera $c$. Otherwise, the cost for a given point is determined by the projected area on the image plane, its deviation from the optical axis due to field curvature~\cite{matsunaga2015field}, and the proximity to the focal plane.
Formally we define the cost as:
\begin{equation}
\begin{aligned}
\kappa^c_s(p)
= {}& w_1 \cdot \min\underbrace{\left(\tfrac{\varepsilon_1\langle p-p_c,\vec{n}_c\rangle^2}{\langle\vec{n}_c,\vec{n}_p\rangle}, 1\right)}_{\hbox{\tiny projected area}} \\
{}+{}& w_2 \cdot \min\underbrace{\left(\tfrac{\|\pi_{\vec{n}_c}(p-p_c)\|}{\varepsilon_2}, 1 \right)}_{\hbox{\tiny optical axis deviation}} \\
{}+{}& w_3 \cdot \underbrace{\left(1 - \mathbbm{1}(p \in V^c_s)\right)}_{\hbox{\tiny proximity to focal plane}}
\label{eqn:cost function}
\end{aligned}
\end{equation}
where
\begin{itemize}  
\item $w_i\in\R^{\geq0}$ is the weight for the $i$-th term,
\item $\varepsilon_i\in\R^{\geq0}$ are thresholding values,
\item $p_c \in \R^3$ is the position of camera $c$,
\item $\vec{n}_p\in S^2$ is the surface normal at $p$,
\item $\vec{n}_c\in S^2$ is the viewing direction of camera $c$,
\item $\pi_{\vec{n}}:\R^3\rightarrow\R^3$ is the projection onto the plane perpendicular to $\vec{n}$,
\item $\mathbbm{1}(\cdot)$ is an indicator function, equal to one if the condition is true and zero otherwise, and
\item $V^c_s\subset\R^3$ is the view frustum of the camera with the near and far clipping planes set to the near and far depth of field limits (shown in \cref{fig:dof}) of camera $c$ with focus distance $s$.
\end{itemize}
In the projected area term, the projection area of a 3D patch is a function of the depth (the distance along the optical axis of the camera) and incidence (the alignment between the viewing direction and the surface normal). Note that we define the surface normals as inward-pointing. $\varepsilon_1$ is a threshold for the projected area so that the projected area term approaches zero for an infinite projected area and goes to 1 for a zero projected area. In the optical axis deviation term, the deviation is defined by the distance between the projection of the point $p$ and the image center on the image plane. $\varepsilon_2$ is a threshold so that the optical axis deviation term equals zero for a point projected onto the image center and approaches 1 for infinite deviation. We scale and clamp individual terms to the range $[0, 1]$ and apply equal weights, $w_i=1/3$, to all terms. A visualization of the cost-determining factors is shown in \cref{fig:criteria}.

\subsubsection{Total Cost}
We define the total focus distance cost, $\mathcal{K}:\R^{|\Camera|}\rightarrow\R^{\geq0}$, by integrating over all points, the minimal pointwise cost over all cameras:
\begin{equation}
\label{eqn:total focal cost}
    \mathcal{K}(\Focus) = \int_{\Mesh} \kappa_\Focus(p) \diff p\quad\hbox{w/}\quad
    \kappa_\Focus(p)=\min_{c\in\Camera}\kappa_{\Focus_c}^c(p)
\end{equation}
The solution is then the set of focus distances minimizing the cost:
\begin{equation}
\label{eqn:optimization}
\Focus = \argmin_{\FFocus\in\R^{|\Camera|}}\left(\mathcal{K}(\FFocus) =\int_\Mesh\kappa_\FFocus(p)\diff p\right).
\end{equation}

\subsection{Expectation-Minimization}
\label{s:expectation_minimization}
Noting that the optimization in \cref{eqn:optimization} can be expressed as a simultaneous optimization over assignments of surface positions to cameras, $\phi:\Mesh\rightarrow\Camera$, and focus distances produces the standard EM problem:
\begin{equation}
(\phi,\Focus) = \argmin_{(\pphi,\FFocus)}\left(\mathcal{K}(\pphi,\FFocus)=\int_\Mesh\kappa_{\FFocus_{\pphi(p)}}^{\pphi(p)}(p)\diff p\right).
\label{eqn:em_optimization}
\end{equation}
(The equivalence follows from the fact that any set of focus distances implicitly defines an assignment of points to cameras, with a point assigned to the camera minimizing the pointwise cost.)

In practice, we approximate the solution using Monte-Carlo integration. Concretely, letting $\Points\subset\Mesh$ be a discrete point-set, we set:
\begin{equation}
(\phi,\Focus)=\argmin_{(\pphi,\FFocus)}\sum_{p\in\Points}\kappa_{\FFocus_{\pphi{p}}}^{\pphi(p)}(p).    
\end{equation}

We use the EM approach for computing the assignment $\phi$ and focus distances $\Focus$ by first initializing the focus distances and then alternately fixing the focus distances and solving for the assignments, and fixing the assignments and solving for the focus distances. (See the supplement for details.)

\subsubsection{Assignment step}
In the assignment step, we solve for the function $\phi: \Points \rightarrow \Camera$ given estimated focus distances $\Focus$. This is done in the standard greedy fashion, assigning a point to the camera minimizing the cost for that point:
\begin{equation}
\label{eqn:assignment}
\phi(p) = \argmin_{c\in\Camera}\kappa_{\Focus_c}^c(p).
\end{equation}

\subsubsection{Minimization step}
In the minimization step, we would like to solve for focus distances $\Focus \in \R^{|\Camera|}$ given the estimated assignment $\phi: \Points\rightarrow \Camera$. As the assignments are fixed, this can be done independently for each camera, with the focus distance being the value minimizing the contribution from the assigned points.
\begin{equation}
\label{eqn:minimization}
\Focus_c = \argmin_{s\in\R} \sum_{p \in\phi^{-1}(c)} \kappa_s^c(p)
\end{equation}
Naively, we discretize the set of possible focus distances into $N$ bins (from the closest to the furthest depth in $\phi^{-1}(c)$ w.r.t. camera $c$) and then find the minimizing focus distance.

\subsection{$k$-view optimization}
\label{subsection: k-view}
Because the minimization step separately considers individual cameras, the EM algorithm may get trapped in a local minima. To mitigate this, we propose an approach inspired by the alpha-beta swap algorithm of Boykov~\textit{et al}.~\cite{boykov2001fast}, using an approach that jointly optimizes the assignment and focus distances for a $k$-tuple of cameras.

Concretely, given an initial assignment $\phi:\Points\rightarrow\Camera$ and given a $k$-tuple of cameras $\mathbf{c}=\{c_1,\ldots,c_k\}\subset\Camera$ we would like to solve the assignment problem for the subset $\phi^{-1}(\mathbf{c})\subset\Points$.

\subsubsection{Partitioning Solution Space}
For a given camera $c$, using the fact that the cost from the projected area and the optical axis deviation are independent of focus distance, this reduces \cref{eqn:minimization} to:
\begin{equation}
\label{eqn:minimization reduced}
\Focus_c = \argmax_{s\in\R}\sum_{p\in\phi^{-1}(c)}\mathbbm{1}(p \in V^c_s)
\end{equation}
That is, the optimal focus distance $\Focus_c$ is the distance at which the largest subset of points assigned to camera $c$ are in its view-frustum. Since the summation in \cref{eqn:minimization reduced} is piecewise constant in $s$, we can find an optimal focus distance by partitioning the range of focus distances into intervals over which the summation is constant. Then, finding the optimal focus distance reduces to finding the interval over which the number of in-frustum points is maximized. We note that the partition of the solution space applies the optimization step in EM. Finally, the optimal focus distance can be set to the mid-point of the interval with maximal count. An illustration can be found in the supplement. 

Similarly, for a $k$-tuple of cameras $\mathbf{c}$, we consider the partitions defined by the cameras, taking their Cartesian product to obtain a partition of the $k$-dimensional space of focus distances associated with $\mathbf{c}$:
\begin{equation}
\mathcal{I}_\mathbf{c}=\mathcal{I}_{c_1}\times\cdots\times\mathcal{I}_{c_k}.    
\end{equation}
As before, this partition has the property that the cost, restricted to $\phi^{-1}(\mathbf{c})$, is constant within each cell.
\subsubsection{Implementation}
Using  $\mathcal{I}_\mathbf{c}$, we perform the joint optimization as follows:
\begin{enumerate}
\item Traversing the cells $\iota\in\mathcal{I}_\mathbf{c}$, of the partition,
\begin{enumerate}
\item We use the mid-point of cell $\iota$ as a candidate focus distances for the $k$ cameras,
\item We compute the optimal assignment of the points in $\phi^{-1}(\mathbf{c})$, given the candidate focus distances given by the midpoint of $\iota$, 
\item We compute the cost given the candidate focus distances and the associated assignment.
\end{enumerate}
\item We replace focus distances for the cameras in $\mathbf{c}$ with the ones given by the midpoint of $\iota$ with the lowest cost.
\end{enumerate}

We note that setting $k=1$, our $k$-view approach can be used to minimize Equation (7), without having to discretely sample the space of focus distances at $N$ locations. This reduces the complexity from $O(nN)$ to $O(n)$ and gives the exact minimum. We use this approach in our implementation of the EM baseline.
For $k > 1$, the joint optimization of both the focus distances and the assignment over the $k$-tuple of cameras allows us to bypass some of the local minima in the optimization landscape. In practice, for one iteration in $k$-view, we partition the cameras into maximally independent sets of $k$-tuples. We then update the focus distances and assignment for each $k$-tuples.

\paragraph{Complexity}
Letting $n=|\phi^{-1}(\mathbf{c})|$ be the number of points assigned to the $k$-tuple of cameras $\mathbf{c}$, the number of cells in $\mathcal{I}_\mathbf{c}$ is $O(n^k)$ and it takes $O(n)$ time to compute the optimal assignment and cost associated to each cell. Thus, the run-time complexity of optimizing over a $k$-tuple is $O(n^{k+1})$.

\section{Results}
\label{section:evaluation}

\subsection{Dataset}
\label{ss: dataset}
We evaluate the methods on the 3DBodyTex dataset \cite{saint20183dbodytex,saint2019bodyfitr} to demonstrate one application in total body photography using the proposed method. The dataset consists of 400 textured 3D meshes of human subjects in various poses and shapes obtained from real scans. The average surface area for the entire dataset is 1,840,527 mm$^{2}$ with a standard deviation of 208,567 mm$^{2}$. We uniformly sample 1K points on each mesh and design a cylindrical camera network around the mesh (\cref{ss:camera configuration} provides more details). Example data with camera networks can be found in the supplement.
\subsection{Parameters}
\label{ss:params and bound}
The thresholds (in \cref{eqn:cost function}) for the projection area and optical axis deviation are $10^{-6}$ mm$^{2}$ and 750 mm in all evaluations. The view frustum of a camera is defined in terms of its aperture (assuming a pinhole camera model) and the distance to the near and far clipping planes, approximated as a function of the focus distance $s$ \cite{ray2002applied}:
\begin{equation}
D_N(s) = \frac{H \cdot s}{H + s - F}
\quad\hbox{and}\quad 
D_F(s) = \frac{H \cdot s}{H - s + F }.
\end{equation}
Here $H=10,000$ mm is the hyperfocal length and $F=50$ mm is the focal length, simulating a DSLR camera with shallow depth of field. We use the intrinsic parameters calibrated from a Canon EOS 90D. Unless otherwise stated, the cameras are placed on a regular $7\times24$ (vertical $\times$ angular) cylindrical grid, sampling at a radius of 750 mm. We use $k=2$ in $k$-view optimization for all the experiments.

\subsection{Evaluation}
\label{s:evaluation}
We evaluate the EM and the $2$-view methods and compare them with two baseline methods considering only single views. For a camera $c\in\Camera$ given its visible point-set $\Points_c \subset \Points$, the baseline methods include setting the focus distance $\Focus_c$ to:
\begin{itemize}
\item \textit{closest}: the closest depth w.r.t. the camera for all points in $\Points_c$ and
\item \textit{avg}: the average depth w.r.t. the camera for all points in $\Points_c$.
\end{itemize}
\cref{t:evaluation} shows the quantitative results. We calculate the average and the standard deviation of the total cost ($\mathcal{K}(\Focus)$) across 400 human meshes for each method. From \cref{t:evaluation}, the EM and the $2$-view not only give us a lower average cost than the baseline methods, but they also provide a smaller standard deviation, showing better robustness of the two methods to variations in shape and pose of the human subjects. Additionally, for the EM and the $2$-view methods, we calculate the change in cost, relative to the baseline methods. The EM and the $2$-view methods reduce the relative cost of the single-view methods by at least $24$\% and $28$\% respectively, corresponding to increasing the in-focus surface area by roughly $1550$ cm$^2$ and $1780$ cm$^2$.

As computing the \textit{global} minimum of the total focus cost is combinatorially hard, we cannot say how close the converged solutions are to being optimal. However, we can estimate a conservative lower bound by assuming all the points in the discrete point-set $\Points$ can always be captured in focus (i.e. setting $w_3=0$ in \cref{eqn:cost function}). The estimated lower bound of the average total cost is 232.15 (recall that the maximum total cost is 1,000). 
\begin{table}
    \centering
    \small
\begin{tabular}{|c|r|r|r|r|} 
\hline
Method & \multicolumn{1}{c|}{\textit{closest}} & \multicolumn{1}{c|}{\textit{avg}} & 
\multicolumn{1}{c|}{EM} & \multicolumn{1}{c|}{2-view}  \\
\hline
\hline
$\mathcal{K}(\Focus)$ & $367.0$ & $352.0$ & $267.5$ & $255.2$\\
\hline
$\sigma$ & $26.5$ & $20.3$ & $15.0$ & $14.5$\\ 
\hline
\hline
$\Delta\mathcal{K}(\Focus)_{EM}$ & $27\%$ & $24\%$ & $0\%$ & $-5\%$ \\
\hline
$\Delta\mathcal{K}(\Focus)_{2-view}$ & $30\%$ & $28\%$ & $5\%$ & $0\%$\\
\hline
\end{tabular}
    \caption{Quantitative evaluation for the proposed methods. The relative difference of the costs is computed by the difference between the target cost and the reference cost, divided by the reference cost.}
    \label{t:evaluation}
\end{table}

\begin{figure*}[thpb]
      \centering
      \includegraphics[width=10cm]{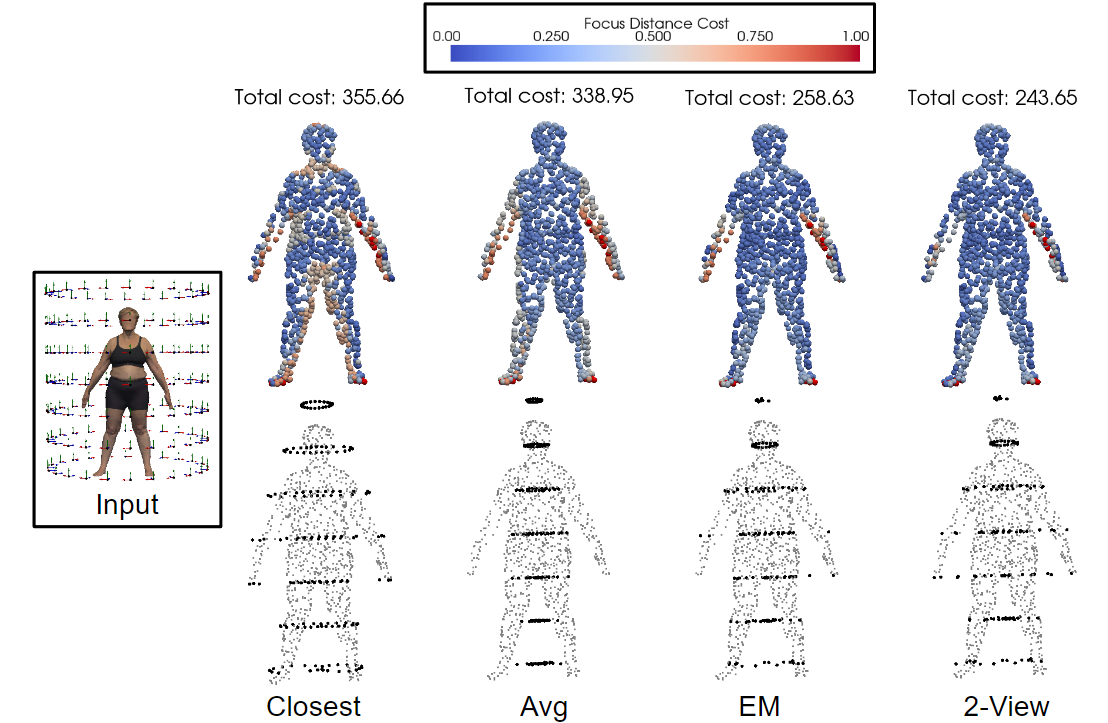}
      \caption{Qualitative comparison of different methods. The first row visualizes the cost of each point. The second row shows the focus distances of cameras as black spheres.  "Closest" and "Avg" are defined in \cref{s:evaluation}.}
      \label{fig:comparison}
   \end{figure*}

\begin{figure*}[thpb]
      \centering
      \includegraphics[width=16cm]{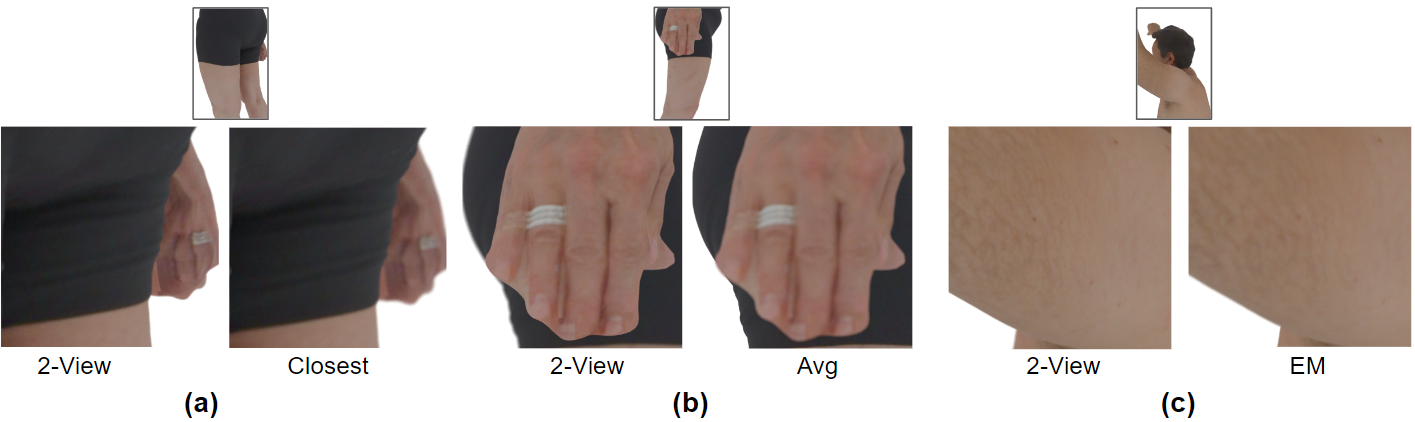}
      \caption{Comparison of image quality from different methods. The first row visualizes the camera view. The second row shows the region where 2-View improves over other methods. (a) compares 2-View with "Closest". Setting the focus distance at the closest depth w.r.t. the camera does not cover the inner part of the right hand in-focus in any view, while 2-View successfully captures it with decent sharpness. (b) compares 2-View with "Avg". Similarly, using the average depth w.r.t. the camera as the focus distance does not handle body parts that are too close to the camera. (c) compares 2-View with "EM". Since 2-View mitigates local minima, we observe more body parts captured in focus using the 2-View method. We note that for regions that are out of focus in the chosen camera view using the 2-View method, they are covered in focus in other views.}
      \label{fig:image comparison}
   \end{figure*}

\cref{fig:comparison} shows the qualitative results of the proposed methods and baseline methods. \cref{fig:image comparison} shows the comparison of the image quality from different methods, with images simulated in Blender\footnote{https://www.blender.org/}. We observe that both EM and $2$-view mitigate the poor imaging of body parts (such as the arms, hands, and inner parts of legs) resulting from a large variation in depth, by carefully selecting the focus distance for each camera. More examples with different poses and shapes can be found in the supplement.
   
\subsection{Sampling Density}
\label{ss:sampling density}
We estimate the integral in \cref{eqn:em_optimization} using a discrete summation. While a larger number of Monte-Carlo samples provides a more robust estimate, it also increases the computational complexity (due to the larger number of partitions in $\mathcal{I}_c$). To better understand the trade-off, we consider variance in the integral estimates for different numbers of samples.

In particular, using different numbers of point samples, $|\mathcal{P}|$, \cref{t:variance} gives the standard deviation ($\sigma$) and running time (in seconds) of an iteration of the $2$-view optimization (computed over 10 different estimates of the integral). Noting that the average total cost (\cref{t:evaluation}) is close to $250$, we believe that the standard deviation of $\sigma=1$ at $|\mathcal{P}|=2^{10}$ represents a stable estimate of the integral. Independently, the table also corroborates the quadratic complexity of the $2$-view algorithm, showing an increase of close to $4\times$ for an increase of $2\times$ in sample count.
\begin{table}
    \centering
\begin{tabular}{|c|c|c|c|c|c|c|c|} 
\hline
$|\mathcal{P}|$ & $2^7$ & $2^8$ & $2^9$ & $2^{10}$ & $2^{11}$ & $2^{12}$ & $2^{13}$  \\
\hline
\hline
$\sigma$ & $2.4$ & $2.6$ & $1.5$ & $1.1$ & $0.9$ & $0.6$ & $0.4$\\
\hline
Time & $3.7$ & $3.8$ & $4.2$ & $6.6$ & $17.9$ & $66.3$ & $259.2$\\ 
\hline
\end{tabular}
    \caption{Stability and computational efficiency of the cost estimate as a function of the number of samples. Stability is measured in terms of the standard deviation, $\sigma$, in the estimate of the integral, taken over 10 samplings. Computational efficiency is measured in seconds.}
    \label{t:variance}
\end{table}

\subsection{Camera Configurations}
\label{ss:camera configuration}
Sampling over a regular cylindrical grid, our camera placement is described by three parameters: The number of angular samples ($\mathbf{a}$), the number of vertical samples ($\mathbf{z}$), and the radius ($\mathbf{r}$). Since the radius ($\mathbf{r}$) should be adjusted based on the quality criteria in \cref{eqn:cost function}, we fixed the radius at 750 mm for all camera configurations.

\begin{figure}
      \centering
      \includegraphics[height=3.5cm]{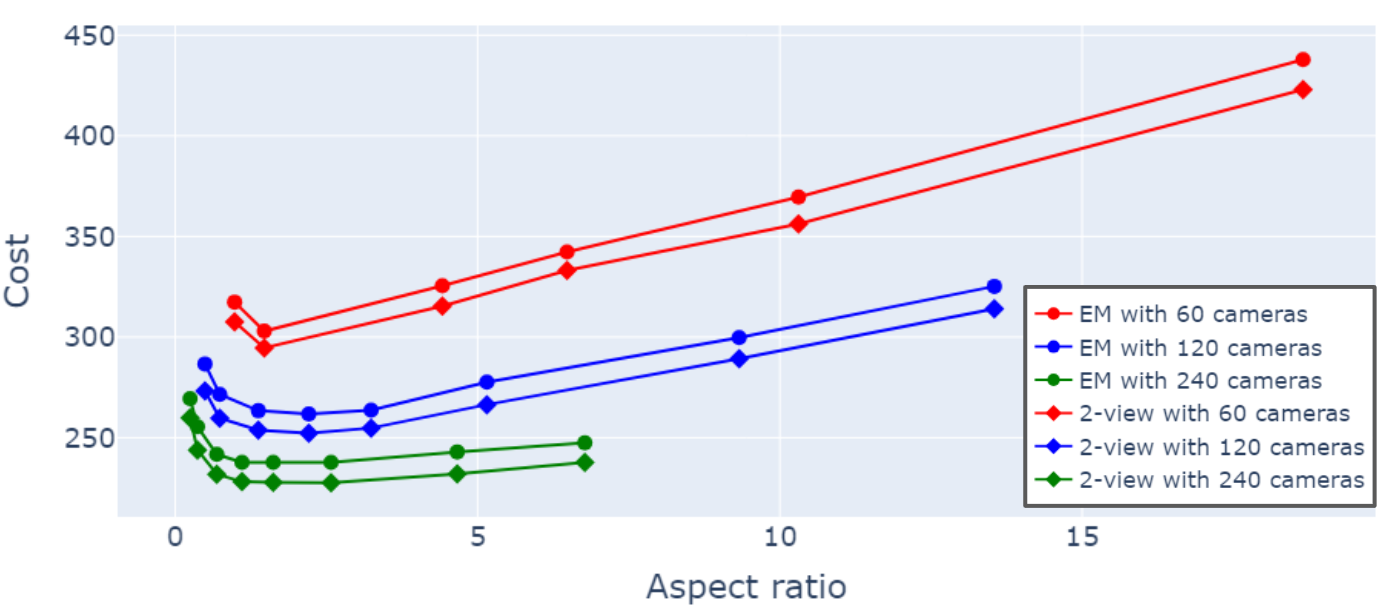}
      \caption{Evaluation of the proposed methods under different number of cameras and different camera configurations. }
      \label{fig:comparison camera configuration}
\end{figure}

\cref{fig:comparison camera configuration} shows the estimated costs for different camera configurations. As we expect the cost to reduce with the number of cameras, we plot the cost against the aspect ratio of the distance between vertical and angular samples. The expected reduction in the cost as a function of the number of cameras is evidenced by the fact that the green curves (240 cameras) are uniformly lower than the blue curves (120 cameras), which are uniformly lower than the red curves (60 cameras). Additionally, we again confirm the benefit of the $2$-view optimization, with the diamond curves ($2$-view) uniformly lower than the circle curves (EM). Finally, we note that regardless of the number of cameras or the optimization technique the minimal cost is attained at an aspect ratio of roughly $3:2$. This matches the fact that the images themselves have a $2:3$ aspect ratio (W/H) in our simulation.

\subsection{Effectiveness of $k$-view Optimization}
\label{ss:effectiveness of kview}
\subsubsection{Overlap within camera tuples}
\begin{figure}[thpb]
      \centering
      \includegraphics[height=5cm]{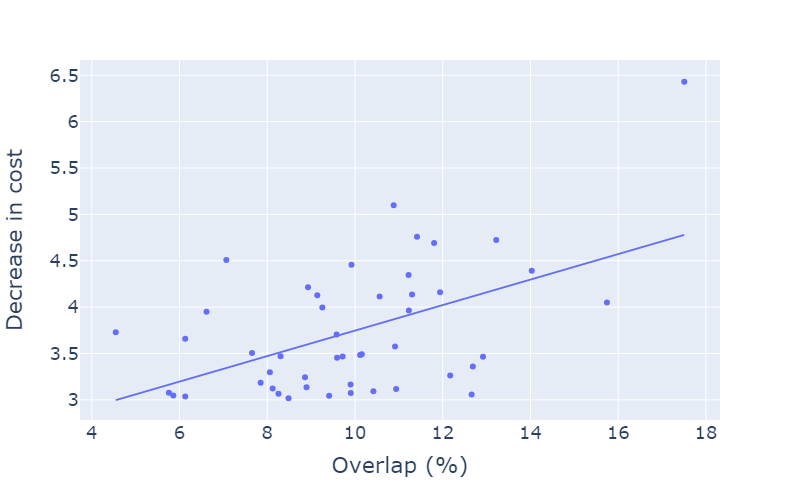}
      \caption{Relationship between the effectiveness of the $2$-view algorithm in various camera pairs and the visibility overlap of individual pairs of cameras.}
      \label{fig:kview in overlaps}
\end{figure}

In general, we expect $2$-view optimization to be more beneficial for camera pairs with more points in common. To assess this, we consider multiple pairs of cameras and for each pair computed the percentage of points shared by the two cameras (overlap), and the decrease in the estimated cost gained by performing $2$-view optimization. \cref{fig:kview in overlaps} visualizes these values as a scatter-plot. The plot shows a clear correlation between the decrease in estimated cost and the percentage of points shared by the two cameras.

\subsubsection{Choosing Camera Tuples}
Given the $k$ views and $|\Camera|$ cameras, there are $\binom{|\Camera|}{k}$ different combinations of $k$-tuples we could consider. However, as discussed above, we expect the joint optimization to be most beneficial when the subset of points visible to all $k$ cameras is large. We take advantage of this by only considering $k$-tuples comprised of cameras that are adjacent to each other on the cylindrical grid. Empirically, we validate that for $k=2$, there is no noticeable benefit ($0.4\%$ relative difference in cost) in considering pairs of cameras that are within a 2-ring neighborhood over using pairs within a 1-ring. (In particular, for $2$-view optimization, the considered camera pairs correspond to the edges of the cylindrical grid.)

\subsubsection{Size of Camera Tuples}
 We also investigate the effect of the size of the tuples on the quality of the $k$-view algorithm. Due to the cost of the optimization, we limit the evaluation to $k=2$ and $k=3$ using 1-ring adjacency in defining the tuples. Measuring the reduction in cost, we find that switching from two views to three provides a little improvement ($1.2\%$ relative difference in cost) but does not validate the significantly increased computational cost ($300\times$ per iteration). 
\section{Discussion}
\label{discussion}
\paragraph{Initialization}
We consider different focus distance initializations and do not observe significant differences, suggesting that the proposed method converges well in practice. As expected, initialization has even less of an effect on the $2$-view algorithm as the $2$-view algorithm is less prone to get trapped in local minima. Furthermore, both EM and $2$-view optimization appear to converge after several iterations. More details can be found in the supplement.
\paragraph{Generalization}
Our implementation of $k$-view optimization uses the fact that the cost function is piecewise constant in the parameter $\Focus\in\R^{|\Camera|}$. The approach generalizes directly to arbitrary cost function by approximating the cost function with a ``piecewise constant'' function, i.e. when the domain over which we are optimizing can be partitioned into cells with constant cost. However, the sampling of the approximated function would play an important role since there is a trade-off between finding an optimal solution evaluated back in the ideal cost function versus the efficiency of the computation.
\paragraph{Assumption of 3D geometry}
Our method assumes knowing the 3D geometry in advance. With the advancement of commodity depth sensors, acquiring an initial 3D mesh is becoming common. For example, \cite{skuka2022extending} proposed a method to extend the depth field of their imaging systems based on the depth map of the scene. In our experiment, we determine the focus distances and evaluate the quality over the same underlying 3D mesh (while sampled differently), we understand that the 3D geometry from depth sensors may be incomplete and noisy in the real world. We would like to investigate its effect in the future. However, how good a 3D geometry of the target can be acquired is beyond the scope of the paper.
\paragraph{Limitations}
The proposed method has several limitations. First, we neglect the issue of focus breathing\cite{rowlands2017physics} -- the change of field-of-view (FOV) during focusing. Therefore, the points captured at the image boundary in the real world may differ in the simulation. However, for close-range photogrammetry in practice, we do not observe a significant effect on the change of FOV when changing the focus distance (lens's focus point) using a constant focal length lens, especially when the range of focus distance is bounded. Additionally, the designed cost accounting for the deviation from optical-axis (in \cref{eqn:cost function}) penalizes points that are assigned far from the image center. Second, our approach assumes a fixed camera configuration where the only degrees of freedom are the focus distances of the individual cameras. In practice, we would also want to consider optimization over a larger space, including camera orientation and zoom. Finally, our method does not apply to objects that deform during the scan. 

\section{Conclusion}
\label{conclusion}
We formulate the multi-view scanning problem when restricted to determining the camera's depth of field. Starting from the standard EM approach, we propose a novel $k$-view optimization method to further improve the quality surface coverage. Empirically in the total body photography, we find that the method can carefully select the focus distance for each camera to handle the poor imaging of body parts resulting from a large variation in depth. The proposed method is robust to different initializations, stable with respect to 1K point sampling, and quickly convergent. Furthermore, while the approach generalizes to $k$-view sampling, we find $2$-view sampling to be sufficient in practice.
As mentioned above while our implementations uses a piecewise constant function, it can readily be extended to a broader class of ``piecewise optimizable'' functions. It would also be straightforward to extend our approach to support weighted optimization, for contexts where it is more important that some subset of the surface be in focus. Although we focus on total body coverage, we believe the algorithms can be useful in other close-range photogrammetry applications.


{\small
\bibliographystyle{ieee_fullname}
\bibliography{egbib}
}

\end{document}